\documentclass{article}
\usepackage{arxiv}

\usepackage[utf8]{inputenc} 
\usepackage[T1]{fontenc}    
\usepackage{hyperref}       
\usepackage{url}            
\usepackage{booktabs}       
\usepackage{amsfonts}       
\usepackage{nicefrac}       
\usepackage{microtype}      
\usepackage{lipsum}

\usepackage[noadjust]{cite}

\usepackage{amsmath}
\usepackage{amssymb}
\usepackage[mathscr]{euscript}
\usepackage[utf8]{inputenc}
\usepackage{flafter}

\newcommand{\dr}{${\Delta R}_{ds(on)}$}

\usepackage{color}
	
\usepackage{algorithm,algorithmicx,algpseudocode}

\usepackage{array}
\usepackage{makecell}

\usepackage{hyperref}
\hypersetup{pdfborderstyle={/S/U/W 1}}

\usepackage{graphicx, subfigure}

\usepackage{tikz}
\usetikzlibrary{matrix,calc}

\begin{document}
	
	
%
%

\title{Real-time Deep Learning at the Edge for Scalable Reliability Modeling of Si-MOSFET Power Electronics Converters\\}


%
%
%

\author{Mohammadreza~Baharani\thanks{\textsuperscript{\textcopyright} 2019 IEEE.  Personal use of this material is permitted.  Permission from IEEE must be obtained for all other uses, in any current or future media, including reprinting/republishing this material for advertising or promotional purposes, creating new collective works, for resale or redistribution to servers or lists, or reuse of any copyrighted component of this work in other works. DOI: \href{https://doi.org/10.1109/JIOT.2019.2896174}{https://doi.org/10.1109/JIOT.2019.2896174}}\\
	Electrical and Computer Engineering Department\\
	The University of North Carolina at Charlotte \\
	Charlotte, NC, 28223 USA\\
\texttt{mbaharan@uncc.edu}\\
\And
Mehrdad~Biglarbegian\\
	Electrical and Computer Engineering Department\\
	The University of North Carolina at Charlotte \\
	Charlotte, NC, 28223 USA\\
	\texttt{mbiglarb@uncc.edu}\\
\And
Babak~Parkhideh\\
Electrical and Computer Engineering Department\\
The University of North Carolina at Charlotte \\
Charlotte, NC, 28223 USA\\
\texttt{bparkhideh@uncc.edu}\\
\And
Hamed~Tabkhi\\
Electrical and Computer Engineering Department\\
The University of North Carolina at Charlotte \\
Charlotte, NC, 28223 USA\\
\texttt{htabkhiv@uncc.edu}
}
\maketitle

%
\begin{abstract}
With the significant growth of advanced high-frequency power converters, on-line monitoring and active reliability assessment of power electronic devices are extremely crucial. This article presents a transformative approach, named Deep Learning Reliability Awareness of Converters at the Edge (Deep RACE), for real-time reliability modeling and prediction of high-frequency MOSFET power electronic converters. Deep RACE offers a holistic solution which comprises algorithm advances, and full system integration (from the cloud down to the edge node) to create a near real-time reliability awareness. On the algorithm side, this paper proposes a deep learning algorithmic solution based on stacked LSTM for collective reliability training and inference across collective MOSFET converters based on device resistance changes. Deep RACE also proposes an integrative edge-to-cloud solution to offer a scalable decentralized devices-specific reliability monitoring, awareness, and modeling. The MOSFET convertors are IoT devices which have been empowered with edge real-time deep learning processing capabilities. The proposed Deep RACE solution has been prototyped and implemented through learning from MOSFET data set provided by NASA. Our experimental results show an average miss prediction of $8.9\%$ over five different devices which is a much higher accuracy compared to well-known classical approaches (Kalman Filter, and Particle Filter). Deep RACE only requires $26ms$ processing time and $1.87W$ computing power on Edge IoT device.
\end{abstract}



%
\section{Introduction}
\label{sec:introduction}

Power electronics systems are essential components of the energy-conversion process. It is expected by 2030, power converters will be used in 80\% of applications in the generation, transmission, distribution, and consumer electronics \cite{tolbert2005power}.  Controllable power semiconductor devices play the most dominant role in the switching power converters. Operating at high current and voltage creates extreme stresses on the power devices, which often makes them the most susceptible components in the energy conversion process. Therefore, understanding, modeling and predicting the reliability models of the power converters are crucial for enabling emerging technologies and future applications such as electric vehicles, smart grids, and renewable energy.

Mathematically formulating and precise understanding of the physical degradation in high-frequency power converters is notoriously difficult, due to the system sophistication and many unknown non-deterministic variables. To solve this problem, a wide range of stochastical diagnostic and prognostics techniques have been proposed to address the reliability issues of a complex system in the design, fabrication, and maintenance process. The evaluation of these processes is beneficial to enable power convertors health management systems and resiliency for useful life estimation and reducing the risk of failures \cite{yang2018design,ma2016new}. Kalman filter and Bayesian calibration are two examples of classical time series modeling and prediction techniques. However, these approaches are often bounded to first-order models in isolation and are not able to bring the collective behavior of many devices with the same underlying physic to create an accurate algorithmic construct. Therefore, their prediction accuracy is very limited. Moreover, they have very limited scalability for emerging advanced technologies \cite{saxena2008metrics, chung2015reliability}. 

Recent advances in deep learning open a new horizon toward smart and autonomous systems. Deep learning offers a scalable data-driven discriminative paradigm to understand, model and predict the behavior of complex systems by extracting the deep collective knowledge.  With the new wave of the Internet-of-Things (IoT) and the feasibility of using the internet almost everywhere, there is a big chance for scalable device-specific real-time monitoring and analysis  by pushing deep learning  and advanced analytic computations from the cloud next to IoT devices (which also called edge computing) \cite{bedi2018review,wang2017predictive}. In particular, the benefits of edge computing are much more pronounced for real-time reliability modeling and prediction of sophisticated physical and engineering systems such as power electronic converters. 

This paper presents a transformative solution, which is called Deep learning Reliability Awareness of Converters at the Edge (Deep RACE)\footnote{The Deep RACE is an open source project and its code is available at \href{https://github.com/TeCSAR-UNCC/Deep\_RACE}{\ttfamily{\textcolor{blue}{ https://github.com/TeCSAR-UNCC/Deep\_RACE}}}.}, for real-time reliability modeling and assessment of power semiconductor devices embedded into a wide range of smart power electronics systems. Deep RACE departures from classical learning and statistical modeling to deep learning based data analytics, combined with full system integration for scalable real-time reliability modeling and assessment. In this regard, it leverages the Long Short-Term Memory (LSTM) networks as a branch of Recurrent Neural Networks (RNN) to aggregate reliability across many power converters with similar underlying physic. Also, It offers real-time online assessment by selectively combining the aggregated training model with device-specific behaviors in the field.

To guarantee real-time scalable requirements, the Deep RACE presents an integrated cloud-edge platform in which, the cloud is responsible to aggregate different device reliability by training the LSTM network, while the inference is done at the edge next to the power devices.  The interference at the edge (on-line) provides real-time feedback of the reliability modeling as well as active control and decision making for device proliferate.  We have trained and implemented the proposed Deep RACE approach for five high-frequency MOSFET power converters. Our results demonstrate the Deep Race improves the misprediction by 1.98x and 1.77x compared to Kalman Filter and Particle Filter, respectively.

To the best of the author’s knowledge, Deep RACE is the first integrative solution for active reliability assessment of the high-frequency power converters based on real-time deep learning analytic. In this context, this paper moves beyond mainstream device modeling and traditional reliability analysis by combining advanced sensing solutions with cutting-edge deep learning and edge computing techniques.  Although this paper primarily focuses on the reliability modeling of high-frequency MOSFETs, the proposed algorithmic construct and system level solution for real-time reliability and predictive maintenance can be extended to a  wide range of semiconductor devices and engineering systems used in power conversion and smart energy systems.

The rest of this article is organized as the following: Section \ref{sec:relatedWork} briefly reviews the previous reliability approaches. Section \ref{sec:Background} provides background on deep learning in particular deep RNN. Section \ref{sec:hybridCondition} presents our proposed Deep RACE approach including LSTM-based machine learning and system-level integration for aggregated training and real-time inference. Section \ref{sec:ExpResults} presents the experimental results including comparison with existing approaches, and finally Section \ref{sec:Conclusion} concludes this article.

%
\section{Related Work}\label{sec:relatedWork}

This section briefly reviews the previous reliability approaches in power electronics, and discusses precursor identifier for power MOSFET degradations.

\subsection{Reliability analysis/prediction in power electronics}

The reliability approaches in power electronics systems have been developed in four broad categories: a) component level, b) damage accumulation, c) data analytics, and d) condition-based predictions. The first approach is not a prognostic-based since they consider the unit-to-unit difference and not the usage history \cite{yang2010condition,degrenne2014review}. The second method offers more accurate tendency for an individual unit by using the accumulation of stress conditions over the time, although it needs experimental observation for the modeling \cite{celaya2012prognostics,musallam2014application}. Data analytics focus on big-data mining for prediction based on the past usage history data and assign a predictive score as opposed to calculating a time to failure events \cite{alghassi2015computationally,heydarzadeh2017bayesian}. Lastly, condition-based prognostic methods rely on potential mode identifications and finding the root of the failure mechanism based on the individual behavior units of failure model physics\cite{gopireddy2015rainflow,dusmez2015accelerated}. 

Several methods have been proposed for mean-life estimations like six sigma, fault tree analysis, state space, and filtering estimations \cite{celaya2012uncertainty,dusmez2016active}. Due to increasing the large volume of data collected from smart devices, using the existing methodologies have some limitations on extracting the hidden patterns. Therefore, the necessity of applying deep learning algorithms for real-time system health monitoring is crucial. Few recent approaches have already demonstrated the significant benefits of deep learning based reliability monitoring and predictive maintenance in other engineering disciplines \cite{doi:10.1111/mice.12359,jiang2017unsupervised,wang2018software}.

\subsection{Precursor identifications in power MOSFET degradation}\label{sub:pre_idn}

In the most comprehensive industry survey-based studies, power semiconductor devices (e.g., MOSFET) are responsible for at least more than 30\% of the failures \cite{wolfgang2007examples,yang2011industry}. The failure mechanisms in power MOSFET can be categorized into two extrinsic and intrinsic subcategories. The extrinsic failures include the transistor packaging issues and mainly summarized as a bond-wire lift, die solder detachment, and contact migration. Most of these studies verified that the bond-wire lift has a severe effect on the device failure over time \cite{kovacevic2010new,anderson2011line}.

To evaluate device long-term reliability, the general approach is conducting an accelerated life test under power/thermal cycles, and continuously monitoring variations in electrical or mechanical parameters. Based on the most acceptable standards for device qualification in the industry, such as AEC-Q101 \cite{AEC}, and the state-of-the-art research, $I_{dss}$, $T_{j}$, $V_{th}$, $R_{th}$, and $R_{ds(on)}$ are the most common parameters for the device degradation tracking \cite{vichare2006prognostics, yang2010condition, song2017failure, celaya2012uncertainty}.  $I_{dss}$, which refers to drain current at zero bias, can be used for early detection of die-level failures, $T_{j}$ shows the device junction temperature and corresponds to thermal runaway failures, $V_{th}$ shows the gate threshold voltage shifting, $R_{th}$ is the thermal resistance of the device and represents device overheating mostly in the package level. Finally, $R_{ds(on)}$ shows the device drain-source resistance, which represents both device degradation in the die and package level where inherently shows the device internal loss.

\begin{figure}[h]
	\centering
	\includegraphics[width=0.6\linewidth,trim= 6 2 2 18,clip]{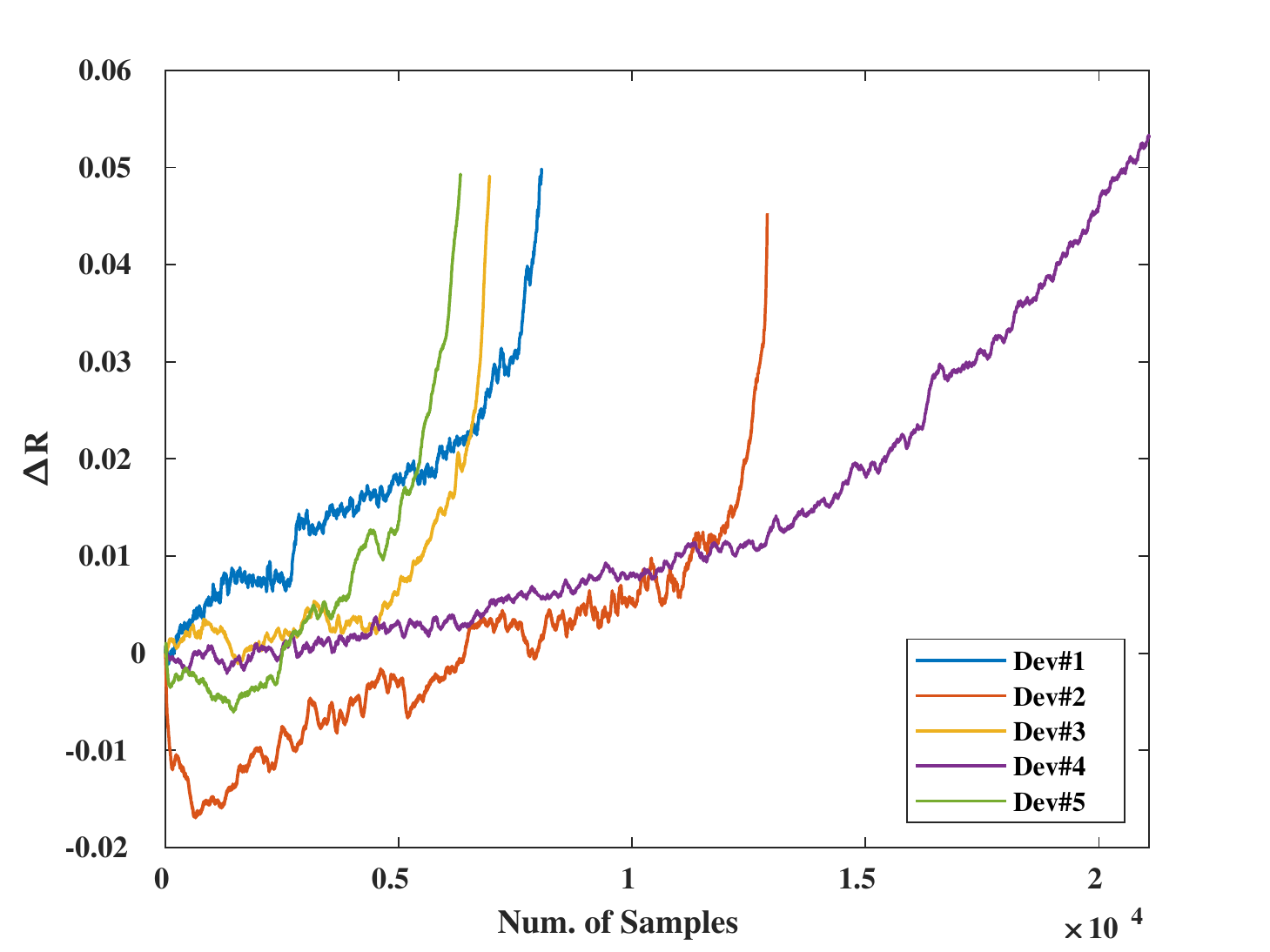}
	\caption{\textbf{MOSFET \dr~Precursor Identifier}: The trajectory $R_{ds(on)}$ for five different MOSFET devices.}
	\label{fig:5devTrajectory}
\end{figure}

Fig. \ref{fig:5devTrajectory} illustrates the changes of $R_{ds(on)}$ over time for five different MOSFET transistors (IRF520NPbf) extracted from the data set provided by NASA \cite{celaya2011prognostics}. Although it may seem that these five transistors share a similar degradation pattern at the first observation, the deterioration pace and detailed behaviors are significantly varied across the devices with similar underlying physics. This is primarily due to diverse workloads, different environmental conditions, and varying manufacturing processes. In the next section, we discuss why the classical approaches are infertile to concurrently model the degradation of these five transistor devices. For the rest of this paper, we also consider these five data set to evaluate the performance of our proposed Deep RACE framework.


%

\section{Motivation and RNN Background} \label{sec:Background}
In this section we explain the limitation of classical approaches for modeling of power device degradation. Then, we continue to elaborate more on vanilla RNN and its problem regarding modeling a complex time series data such as MOSFET \dr~precursor.
 
\subsection{Limitation of Classical Approaches}
In the reliability-based prediction methods, the frequency of component failure is predicted based on a statistical model derived from acquired data in a laboratory environment or historical component usage if available. In a strict sense, these methods cannot be considered as prognostic methods due to unique unit-to-unit conditions and their specific usage history. Alongside, although theoretical approaches such as physics-of-failure \cite{wang2013toward} analysis are applied to identify the root of failure and drive the reliability model, these methods result in significant errors and are not applicable for unit-to-unit scenarios because of the complexity of power electronics systems and their operating conditions \cite{celaya2012uncertainty}.

The other data-driven approaches, such as Kalman filter \cite{dusmez2016remaining} and Bayesian calibration \cite{celaya2011prognostics, heydarzadeh2017bayesian}, are predictive analytics which mostly focus on identifying the correlation of the experimental results, and the estimation of unknown variables and parameters. These techniques require an accurate failure model of a system to estimate the unknown mathematical parameters associated with a specific failure test; however, for new technologies, these methods cannot be effective due to the lack of precise failure model in the component as well as system level \cite{dusmez2015accelerated}. There is a high demand for solutions that address the algorithmic and system-level challenges for real-time scalable monitoring, modeling, and estimation of degradation behavior in power electronic transistors.

\subsection{Recurrent Neural Networks}
In contrast to classical approaches, this article proposes a holistic IoT system for hybrid condition-based prediction model which assesses the behavior of individual transistors based on both their usage history as inferred from sensed data and expected future load profiles. To achieve this, a data-driven model based on deep Recurrent Neural Network (RNN) is utilized at the edge, i.e. converter, for real-time device-specific reliability prediction and modeling; while the cloud infrastructure performs high-level metadata aggregation and analysis across many devices. At the time of Artificial Intelligence (AI) big-bang, we took advantages of a prominent model of RNN named LSTM to predict the transistor degradation. In our solution, the sensed resistance, and environment conditions will be sent to the cloud-side of the proposed framework to train and update the LSTM network models. Therefore, the models will be updated based on the current device operating condition.

\begin{figure}[!t]
	\centering
	\includegraphics[width=3.6in,trim= 40 3 6 15,clip]{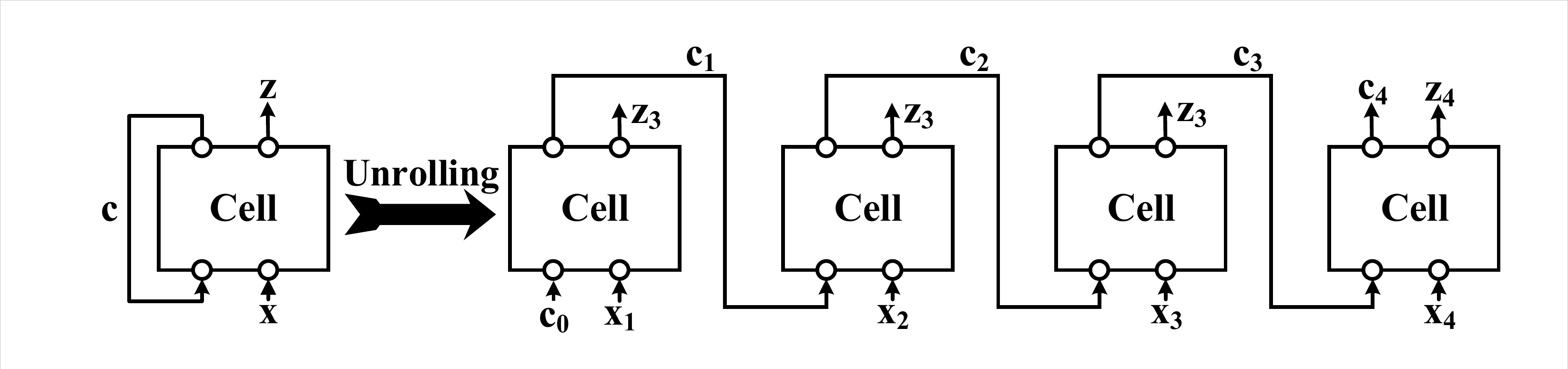}
	\caption{\textbf{Recurrent Neural Networks}: The schematic of standard RNN cell and its unrolling version for four input time sequence}
	\label{fig:rnnUnroll}
\end{figure}

The RNNs are a branch of neural networks specialized for analyzing and modeling complex time series. Following deep learning paradigm, RNNs need a fairly large dataset for training. The are very popular approach for natural language processing and object tracking over frames sequence. In a formalized RNN, sequence of data is notated by $X=[\begin{matrix}x_1&x_2&...&x_\tau\\\end{matrix}]$ where $\tau$ is the number of input sequences. Fig. ~\ref{fig:rnnUnroll} formulates an RNN computation node, i.e.  neuron, and its unrolling recurrent computation when $\tau=4$. A neuron passes the information from the past to the current time by sharing the information and updating its cell state, $c$; therefore, this sharing process enables RNN to model a behavior of a time sequence. In a standard RNN cell with given input $X$, the cell output $Z=[\begin{matrix}z_1&z_2&...&z_\tau\\\end{matrix}]$ is computed by Equations \ref{eq:weightedInput}-\ref{eq:RNNoutput}. In these equations, $\zeta(\cdot)$ and $\xi(\cdot)$ are nonlinear activation functions, $W_i$ is input weight, $W_c$ is the state cell weight, $W_o$ is the output weight, and $b_i$, $b_o$ are biases for input and output values, respectively. 

\begin{gather} 
i_t=W_ix_t+W_cc_{t-1}+b_i \label{eq:weightedInput}, \\
c_t=\zeta(i_t) \label{eq:hiddenLayer}, \\
o_t=W_oc_t+b_o \label{eq:weightedOutput}, \\
z_t=\xi(o_t) \label{eq:RNNoutput}.
\end{gather}

Knowing $Y=[\begin{matrix}y_1&y_2&...&y_\tau\\\end{matrix}]$ as the referenced output, the loss function is defined by (\ref{eq:lossFunction}). In (\ref{eq:lossFunction}), $\mathscr{L}(z_t,y_t)$ can be considered as the squared error of a regression function or cross-entropy for the sake of classification.  The ultimate goal of RNN learning is to minimize the cost function. This goal can be formalized by (\ref{eq:Optimization}), which minimizes the introduced loss function by altering $\theta$, where $\theta$ is a network vector model which is described as: $\theta=[\begin{matrix}W_i&W_c&W_o&b_i&b_o&c_0\\\end{matrix}]$.

\begin{gather} 
\mathscr{ L}(Z,Y)=\sum_{t=1}^{\tau}\mathscr{L}(z_t,y_t)\label{eq:lossFunction},\\
\underset{\theta}{\mathrm{argmin}}{\mathscr{~L}(z(\theta),Y)}\label{eq:Optimization}.
\end{gather}

Backpropagation through time (BPTT) is the mainstream approach to extract the proper weight factors of the RNNs to minimize the cost function across the trained data \cite{werbos1990backpropagation,rumelhart1986learning}. For the vanilla RNN, a major issues associated with BPTT is the losing of cell sensitivity to the earliest inputs due to the chain of partial derivation. This phenomena is known as vanishing gradient problem and eventually prevents the network reaches to the earliest states in deep RNN \cite{hochreiter2001gradient}. More sophisticated versions of  RNN network, such as LSTM cells, are proposed to address the sensitivity lost problem in the basic RNN networks. Based on that, in the next section, we describe our proposed reliability model based on the LSTM networks.

%

\section{Deep Learning Reliability Awareness of Converters at the Edge (Deep RACE)}\label{sec:hybridCondition}

This section presents Deep RACE as an integrative framework of online real-time reliability awareness and modeling for power electronic devices.  Deep RACE has two major aspects: (1) algorithmic principles for modeling the device reliability, and (2) system-level integration for real-time scalable reliability assessment. On the algorithm side, Deep RACE uses one of the major derivatives of RNN, called LSTM for aggregated training and device specific inference. On the system side, Deep RACE proposes an IoT-based edge-cloud computation platform which pushes the proposed LSTM-based reliability inference next to the individual power converters. 
In the following, we provide an in-depth explanation of both aspects.

\subsection{Algorithmic Constructs for Device Reliability Modeling}
In this part, we first introduce the basics of LSTM cells for reliability modeling of power devices, and we continue to present our proposed reliability modeling network constructed out of the basic LSTM cells. 

\subsubsection{Long Short-Term Memory Cells}\label{sec:lstm}
For reliability modeling of power electronic converters, we propose using LSTM cells. In a nutshell, different deep neural networks recognize the patterns in two forms of spatial and temporal pattern depending on their structure. As an instance, Convolutional Neural Networks (CNN) are engineered in a form that they can distinguish spatial pattern, e.g. a dog or a face, existed in a picture. In the other hand, sequence data such as natural language data and time series, e.g. stock index have a temporal pattern that should be processed during a sequence of time. For our case, since we try to model \dr~data and it is intrinsically a time series, we leverage LSTM cells as the prominent version of RNN. The benefits of LSTM cells are in using the guided gates for selectivity remembering both short and long-term behaviors across many time series. The LSTM uses a subset of the cyclical node inside its cell known as ``memory'' in order to calculate the output based on the current input and its past status \cite{hochreiter1997long}. The LSTM selective memory sensitivity at large scale offers a systematic approach for reliability modeling of power electronic devices.

\begin{figure}[h]
	\centering
	\includegraphics[width=0.7\linewidth, trim= 5 5 5 35,clip]{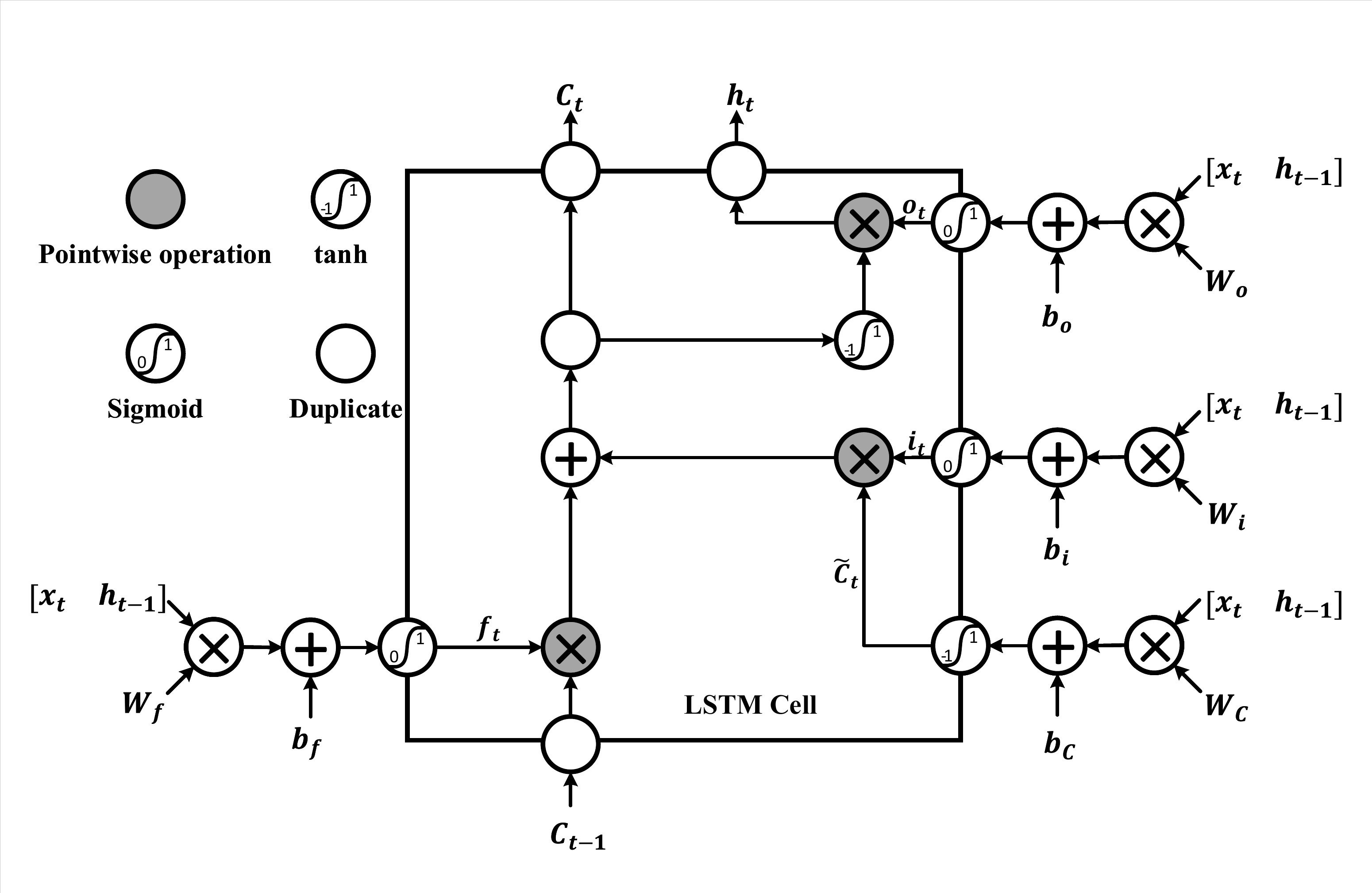}
	\caption{\textbf{A single LSTM cell}: Inside of an LSTM cell consisting of three gates and the state of the cell is preserved by variable $c_{t}$.}
	\label{fig:LSTM}
\end{figure}

The LSTM cell contains three vectorized sigmoid ($\sigma$) functions, which each individual function operates as a gate and controls the flow of information passing through the cell --- the input, output, and forget gates. Fig. ~\ref{fig:LSTM} presents the internal structure of an LSTM cell. Each gate maps its input to $S=\{s_i | s_i \in [0, 1]\}$, where zero is a closed gate and one means the gate is open. Moreover, the cell state (memory) is preserved by $C$ as a candidate. The information of new candidates, which should be stored in the cell state, represented as $\tilde{c}$.
\vspace{-0.6cm}

\begin{gather} 
i_t=\sigma(W_iv_t+b_i) \label{eq:lstmInput}, \\
f_t=\sigma(W_fv_t+b_f)\label{eq:lstmForget}, \\
o_t=\sigma(W_ov_t+b_o) \label{eq:lstmOutput}, \\
\tilde{c}_t=tanh(W_cv_t+b_c) \label{eq:lstmNewCandidate}, \\
c_t=f_t \odot c_{t-1} + i_t \odot \tilde{c}_t \label{eq:lstmCandidate},\\
h_t=o_t \odot tanh(c_{t}) \label{eq:lstmH}.
\end{gather}

With respect to LSTM cell visualization in Fig. ~\ref{fig:LSTM}, Equations \ref{eq:lstmInput}-\ref{eq:lstmH} formulates the correlation between input and output per LSTM cell. In the equations, $v_t=[\begin{matrix}x_t&h_{t-1}\\\end{matrix}]$, $\odot$ is Hadamard or element-wise matrix product, and $\theta=[\begin{matrix}W_i&W_o&W_f&W_c&b_i&b_o&b_f&b_c&c_0\\\end{matrix}]$ is the network model that should be trained. The input gate (\ref{eq:lstmInput}) decides what portion of current input with which degree should be stored in cell memory, while forget gate (\ref{eq:lstmForget}) chooses which portion of memory should be erased. In the other hand, new information, $\tilde{c}$, will be mined by (\ref{eq:lstmNewCandidate}), and the cell memory will be updated by (\ref{eq:lstmCandidate}). Moreover, the output gate (\ref{eq:lstmOutput}) decides which part of cell memory should affect the LSTM output at time $t$, and finally the LSTM output value is calculated in (\ref{eq:lstmH}).

\subsubsection{LSTM-Based Device Training Network}
LSTM-based neural network needs to be designed to properly reflect \dr~ propagation with enough depth to capture complex power convertors behaviors, and thus learning deep behavioral patterns across many devices. A deep LSTM network should have sufficient depth to build up the progressive pattern recognition of sequential data in both coarse and fine grain directions. 

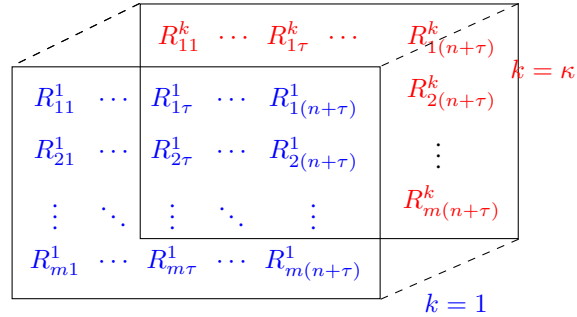
\begin{figure}[h]
	\centering
\begin{tikzpicture}
\matrix (mA) [draw,matrix of math nodes]
{
	\textcolor{red}{R_{11}^k} & \textcolor{red}{\cdots} & \textcolor{red}{{R_{1{\tau}}^k}} & \textcolor{red}{\cdots} & \textcolor{red}{{~~~~R_{1({n+\tau})}^k}} \\
	[] & [] & [] & [] & \textcolor{red}{{~~~~R_{2({n+\tau})}^k}} \\
	[] & [] & [] & [] & \vdots  \\
	[] & [] & [] & [] & \textcolor{red}{{~~~~R_{m({n+\tau})}^k}}  \\
};
\matrix (mC) [draw,matrix of math nodes] at ($(mA.south west)+(0.75,0.75)$)
{
	\textcolor{blue}{{R_{11}^1}} & \textcolor{blue}{\cdots} & \textcolor{blue}{{R_{1{\tau}}^1}} & \textcolor{blue}{\cdots} & \textcolor{blue}{{R_{1({n+\tau})}^1}} \\
	\textcolor{blue}{{R_{21}^1}} & \textcolor{blue}{\cdots} & \textcolor{blue}{{R_{2{\tau}}^1}} & \textcolor{blue}{\cdots} & \textcolor{blue}{{R_{2({n+\tau})}^1}} \\
	\textcolor{blue}{\vdots}	 & \textcolor{blue}{\ddots} &	\textcolor{blue}{\vdots}	   & \textcolor{blue}{\ddots} & \textcolor{blue}{\vdots}			\\
	\textcolor{blue}{{R_{m1}^1}} & \textcolor{blue}{\cdots} & \textcolor{blue}{{R_{m{\tau}}^1}} & \textcolor{blue}{\cdots} & \textcolor{blue}{{R_{m({n+\tau})}^1}} \\
};
\draw[dashed](mA.north east)--(mC.north east) node [near end, below right] {~~~~~~~~~~~~~\textcolor{red}{$k=\kappa$}};;
\draw[dashed](mA.north west)--(mC.north west);
\draw[dashed](mA.north west)--(mC.north west);
\draw[dashed](mA.south east)--(mC.south east) node [near end, below right] {\textcolor{blue}{$k=1$}};

\end{tikzpicture}
	\caption{\textbf{Batch tensor configuration}: Three dimensional batch tensor with a characterized vector ${R_{mt}^k}$.}
	\label{fig_new}
\end{figure}

As it mentioned in Section \ref{sub:pre_idn}, we consider the trajectory resistance of drain-source of power MOSFET during ON time (i.e. \dr) as the precursor of device failure degradation. As \dr~is intrinsically a time series, we design the model by using deep LSTM network, where the training is developed by aggregating data from different devices having the same technology. A batch of samples is created for each training iteration. Therefore, for predicting the next $n$ samples of \dr~based on the provided last input sequence ($\tau$), the batch should consist of \dr~with the size of ($\tau+n$). Fig.~\ref{fig_new} presents the batch tensor configuration per each iteration. The dimension of vector ${R_{mt}^k}$ is characterized based on the \emph{input size} shown by $k$, where $m$ is the available devices for training, and $t$ is the sequence. In order to prevent any sort of biases in training the LSTM, we selected randomly a vector sequence with the size of ($\tau+n$) from the \dr~ samples per each device. Increasing the number of devices, ($m$), per each batch helps the network to generalize the modeling of complex degradation properly, which results in higher accuracy of the predicted trajectory.

For designing of deep LSTM network, we need to also consider \emph{number of hidden layers}. The number of hidden layer is the dimension of vectors generated in equations (\ref{eq:lstmInput})-(\ref{eq:lstmH}). The vector size can be changed by altering the weights and biases tensor shape defined in equations (\ref{eq:lstmInput})-(\ref{eq:lstmNewCandidate}). Increasing the hidden layer is interpreted as increasing the ``memory'' size of the LSTM and its capacity to learn existing complex pattern in a signal. 

\begin{figure}[h]
	\centering
	\includegraphics[width=0.60\linewidth,trim= 20 20 20 20,clip]{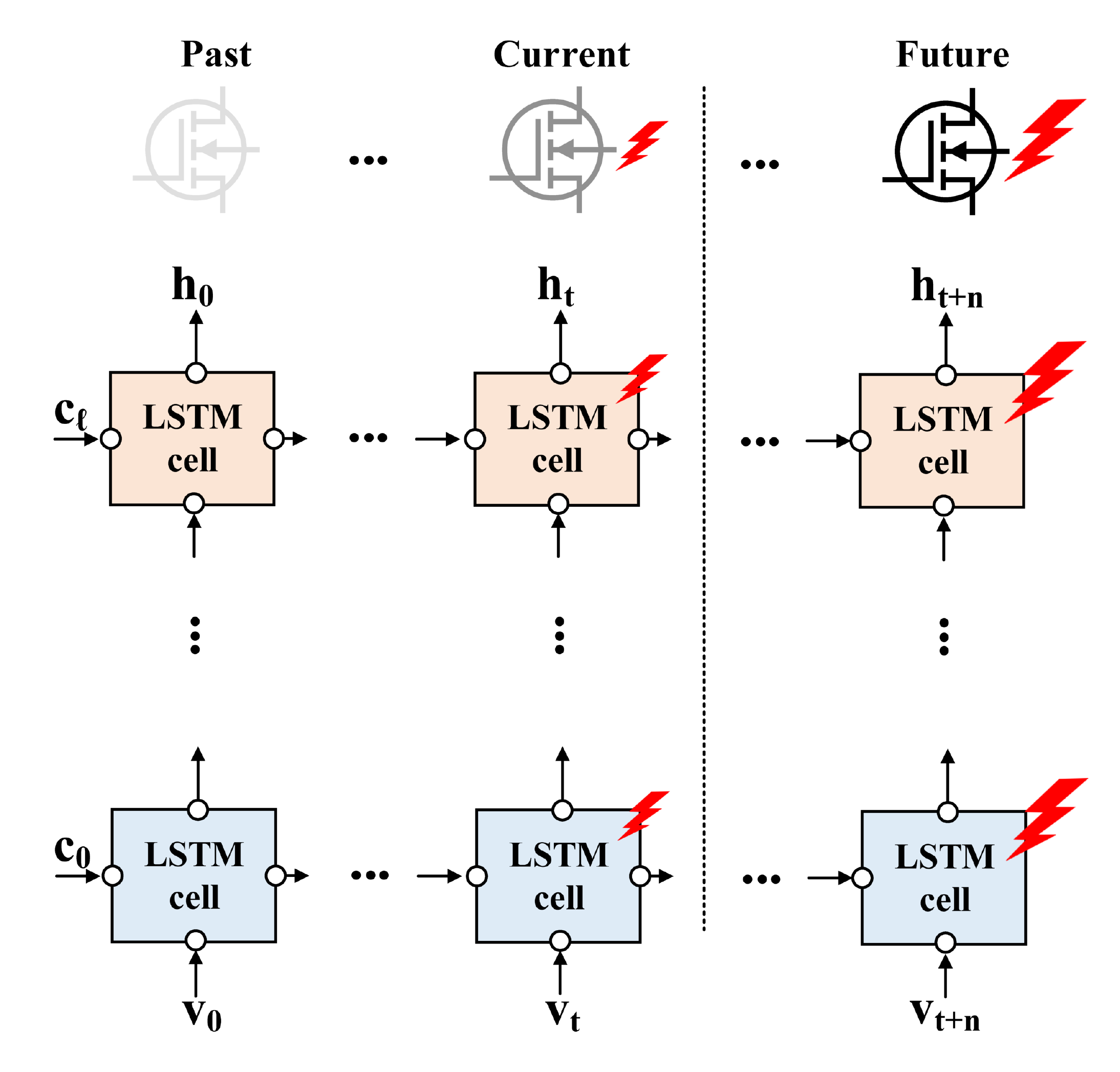}
	\caption{\textbf{The stacked LSTM}: An unrolled LSTM cell predicts the next $n$ samples of \dr~based on last sensed data.}
	\label{fig:LSTM1}
\end{figure}

At the same time, building a very deep LSTM network is not a viable solution due to lack of large data-set to trained all LSTM cells once at the same time. The BPTT often failed to train a large flat LSTMs network without any patriarchy. As the result, the alternative solution is stacking multiple LSTM networks (which called Stacked LSTM) to create more complex and deeper network with offering hierarchical modular layers over a very deep flat network. In our proposed network, we suggest developing a stacked LSTM to generalize the behavior complexity of power electronics convertor without the need for a very large data-set.  As the result, \emph{number of stacked layers} is the other parameter to increase the complexity of the LSTM network. Here, by stacking the cell up together in a way the $h_t$ of one cell is used as an input to its top adjacent cells.  Fig.~\ref{fig:LSTM1} shows an architecture of the deep LSTM network with the stacked layer size of $\boldsymbol\ell$.

\begin{algorithm*}[h!]
	\caption{Training the deep LSTM network}
	\label{alg:training}
	\begin{algorithmic}[1]
		\Require $X_{training}, Y_{training}, X_{test}, Y_{test}, \tau, m, n, \ell, k, \epsilon,
		e_{th}, it_{max}$
		\Ensure $\theta_{\lambda}; 1\leq \lambda \leq \ell, \theta_d$ \Comment{Network models}
		\State $computation\_graph \gets$ LSTM($hidden\_layer, \ell$)
		\State $error$ $\gets \infty$ \Comment{Initialize the test error}
		\State $min\_error$ $\gets \infty$ \Comment{Saves the minimum error test}
		\State  $\jmath \gets$ 0
		\State init\_rand($\theta_{\lambda}$ \textbf{for} $\lambda$ in $[1...\ell]$) \Comment{Initialize LSTM network models from truncated normal distributions}
		\State init\_rand($\theta_d$)
		\While{($\jmath \leq it_{max}$) \textbf{or} ($error \geq e_{th}$)}
		\State $X_{batch}, Y_{batch}$ $\gets$ generate\_data($X_{training}$, $Y_{training}$, $m$, $n$, $\tau$, $k$)
		\State \dr $\gets$ inference($computation\_graph, X_{batch}$, [$\theta_{\lambda}$ \textbf{for} $\lambda$ in $[1...\ell]$], $\theta_d$) \Comment{Predicted \dr~for training}
		\State $error_{training} \gets \mathscr{ L}({\Delta R}_{ds(on)},Y_{batch})$
		\State [$\theta_{\lambda}$ \textbf{for} $\lambda$ in $[1...\ell]$], $\theta_d$ $\gets$ optimizer($computation\_graph, error_{training}$, [$\theta_{\lambda}$ \textbf{for} $\lambda$ in $[1...\ell]$], $\theta_d$) \Comment{Updating the net. models}
		\State $x_{test}, y_{test}$ $\gets$ generate\_data($X_{test}$, $Y_{test}$, $m$, $n$, $\tau$, $k$)
		\State \dr $\gets$ inference($computation\_graph,x_{test}$, [$\theta_{\lambda}$ \textbf{for} $\lambda$ in $[1...\ell]$], $\theta_d$) \Comment{Predicted \dr~for test}
		\State $error$ $\gets \mathscr{ L}({\Delta R}_{ds(on)},Y_{test})$
		\If{($min\_error > error$)}
		\State $min\_error \gets error$
		\State save($\theta_{\lambda}$ \textbf{for} $\lambda$ in $[1...\ell]$, $\theta_d$)\Comment{Save the network model with lowest error}
		\EndIf
		\State $ \jmath \gets \jmath + 1$
		\EndWhile
	\end{algorithmic}
\end{algorithm*}

\begin{figure}[h]
	\centering
	\includegraphics[width=0.60\textwidth,trim= 20 20 20 20,clip]{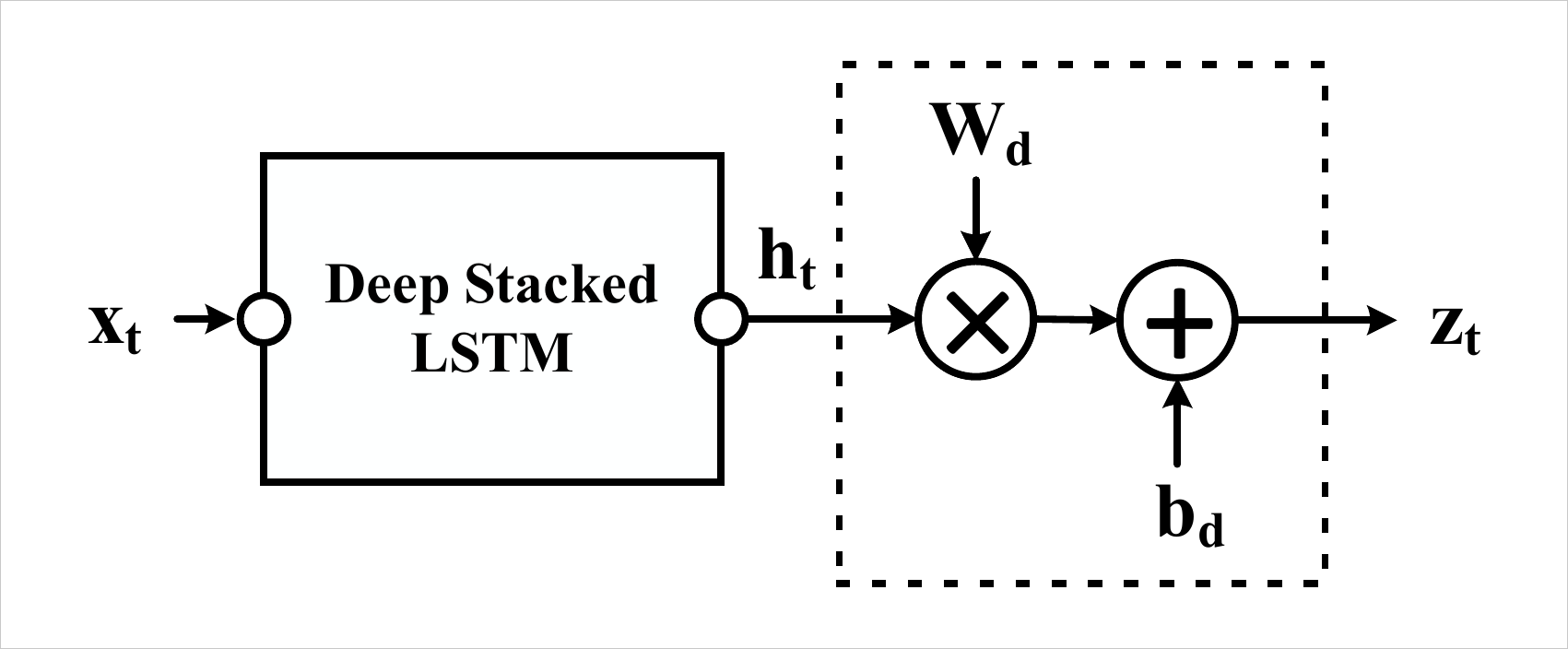}
	\caption{\textbf{The proposed deep LSTM network model}: A dense layer is added to the deep stacked LSTM to map $h_t$ to on-line measured \dr~ at time $t$.}
	\label{fig:wSum}
\end{figure}

Since the output vector $h_t \in [-1, 1]$, we need to de-normalize the deep LSTM network output to actual system measurement. Therefore, a dense layer is added to the output of stacked LSTM to map $h_t$ to the predicted \dr~at the time $t$ as shown in Fig.~\ref{fig:wSum}. Based on modified deep LSTM structure, the network models are described as:  $\theta_{\lambda}=[\begin{matrix}W_{\lambda_i}&W_{\lambda_o}&W_{\lambda_f}&W_{\lambda_c}&b_{\lambda_i}&b_{\lambda_o}&b_{\lambda_f}&b_{\lambda_c}&c_{\lambda_0}\\\end{matrix}], 1\leq \lambda \leq \boldsymbol\ell$, and $\theta_{d}=[\begin{matrix}W_{d}&b_{d}\\\end{matrix}]$, where $\boldsymbol\ell$ is the stacked layer size. Each LSTM cell will be trained at the cloud, and network model will be transferred to the edge next to the transistor device for real-time prediction. Acquiring proper values for the LSTM network parameters can be done by exploring the design space in regard to system constraints (e.g., system accuracy, processing time, and power consumption) \cite{Baharani:2014:HDS:2665611.2665834}.

\subsection{Proposed IoT Framework}

In this subsection, we explained the system integration to realize the proposed LSTM reliability modeling constructs through an IoT-based cloud-edge platform. 

\subsubsection{Data training and batch aggregation on the cloud}
\begin{figure}[h]
	\centering
	\includegraphics[width=0.6\linewidth,trim= 10 10 10 10,clip]{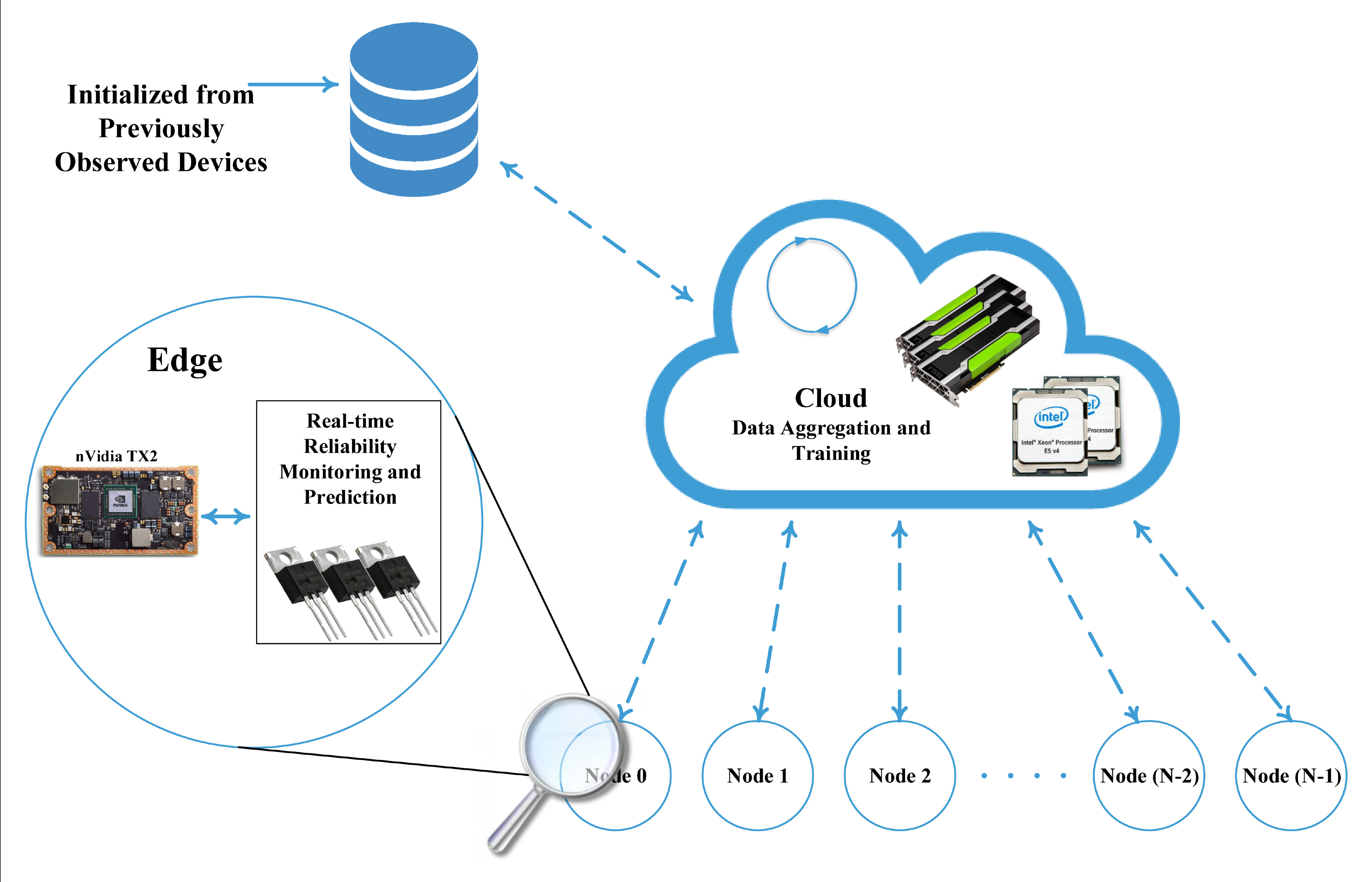}
	\caption{\textbf{The Deep RACE Framework}: The proposed solution accumulates the knowledge of power transistor degradation model on the cloud-side by training the LSTM network, while real-time prediction and inference is accomplished on the edge side.}
	\label{fig:deepRace}
\end{figure}

One key aspect of Deep RACE is an integrative cloud-edge system for scalable real-time reliability monitoring, assessment, and prediction. Fig. \ref{fig:deepRace} shows the system architecture of Deep RACE based on the cloud-edge solution - training on the cloud and real-time sensing and inference for reliability monitoring and health assessment on the edge nodes. The cloud is the centralized computing node for data aggregation and training of proposed LSTM algorithm across many power electronic transistors with similar underlying physics. The cloud stores the initial sampled dataset (e.g., voltage, current, and device temperature) collected from devices under stress test for initial training. At the same time, it continuously receives new information and sample data from running devices for improving the accuracy of reliability modeling and prediction. The edge nodes are IoT devices which use local computing power to perform the real-time reliability monitoring, and prediction (deep learning inference). The edge nodes rely on the pre-trained models that have created during the training phase on the cloud.

In this context, the edge nodes and cloud continuously interact and exchange information. The Edge nodes, while performing real-time reliability assessment, collect and update the cloud with the properties of the transistors for future training. The cloud also updates edge nodes with new reliability models (LSTM models) to increase the confidence interval of the prediction, and estimate the remaining useful life of the devices. To increase the confidence interval of the power MOSFET devices and the system operation, a predefined threshold ${\Delta R}_t$ can be defined based on system requirement. Once the error of predicted device resistance \dr~ is greater than the ${\Delta R}_t$, the network models will be automatically updated through training the network on the cloud server.

\begin{figure}[h]
	\centering
	\includegraphics[width=0.85\textwidth,trim= 14 10 10 10,clip]{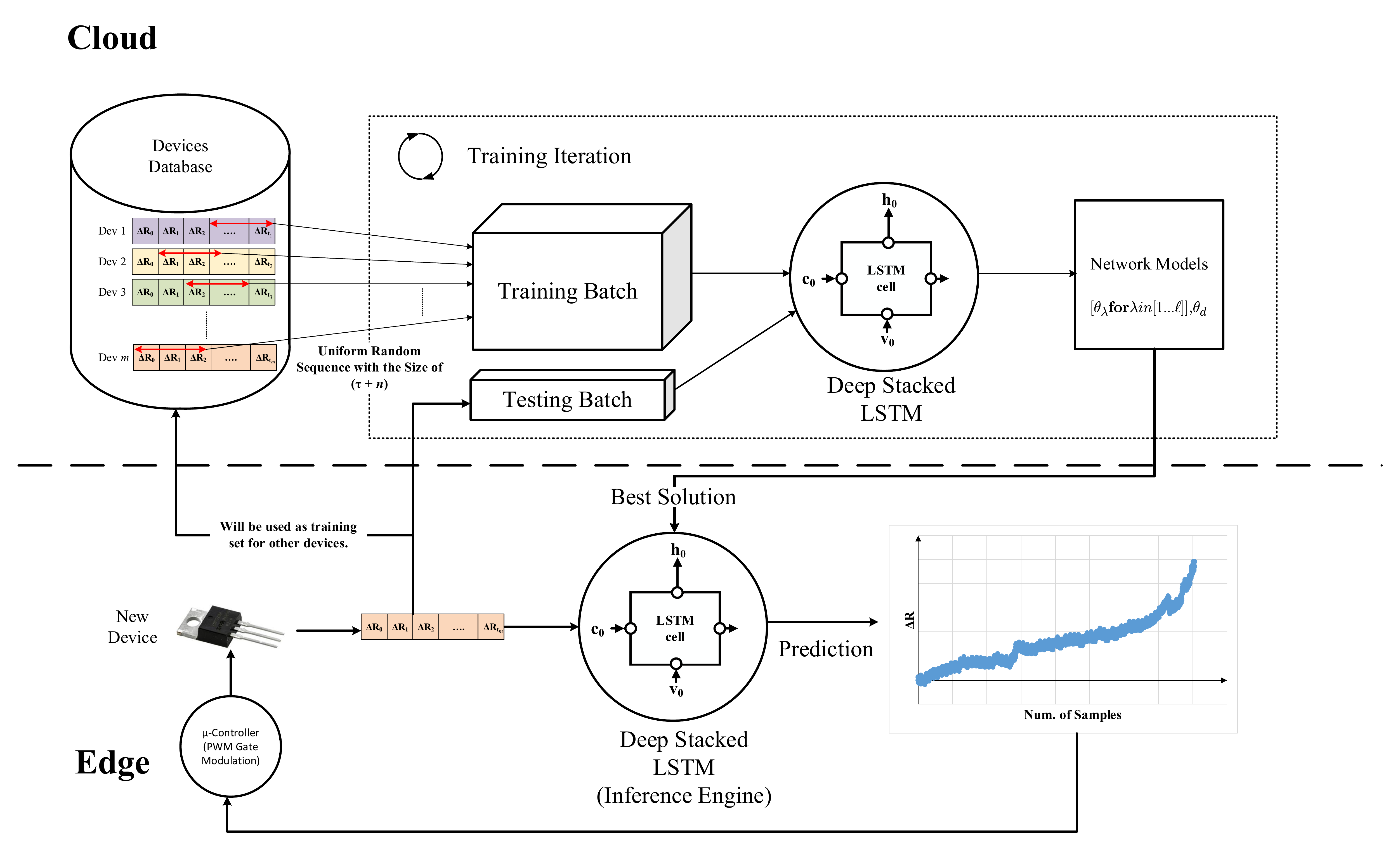}
	\caption{\textbf{The scalability of Deep RACE}: As new edge is added to the framework, its extracted \dr~ is transfered to the cloud to be used both as training data set (for other devices) and test data set to prevent over fitting problem.}
	\label{fig:deepRaceAlg}
\end{figure}

Algorithm \ref{alg:training} describes the training procedure on the cloud side. The $computaion\_graph$ is the deep LSTM network structure. At first, the network models are initialized from a truncated normal distribution. $X_{batch}, Y_{batch}$ are generated based on the input size, input sequence, output sequence, and the number of devices. Next, $computation\_graph$ predicts the \dr~according to its current network models and generated training batch. The models will be updated through the back-propagating error from the output of $computaion\_graph$ to its inputs. 

During training phase of our proposed LSTM algorithm, and in general deep learning models, one major source of error would be over-fitting. In order to prevent the over-fitting problem, we have created a test batch ($x_{test}$, $y_{test}$) to predict \dr~ of the test device based on the updated network model. Next, the test batch $error$ will be compared against the previous error values and if it has the minimum value, the network model will be saved.

\subsubsection{Real-time edge analysis}
The edge converter has its own local controller equipped with an embedded SoC for the purpose of predicting the power transistor degradation. The $\mu$-controller unit is responsible of modulating the gate signals for the power converter control, and continuous monitoring of the voltage, current, and temperature of power converters. The sampled data also will be transferred to the cloud , which performs the reliability analysis (training phase) for each edge node. The SoC of the edge runs inference section of the deep LSTM, equations (\ref{eq:lstmInput})-(\ref{eq:lstmH}), and estimates the trajectory device resistance, \dr, based on the received trained network models from the cloud. Based on the predicted \dr~, controller can leverage load-sharing \cite{hu2013interleaving} method through the system level control in modular converters or cascaded architectures in many applications such as distribution generation systems, data centers, and the electric vehicles in order to decrease the degradation pace until the appropriate action is taken.

Fig.~\ref{fig:deepRaceAlg} visualizes the scalability of Deep RACE when a new edge is added to the framework. At first, the edge node sends the voltage, current, and temperature samples to the cloud-side. The samples will be used for two purposes: (1) as training sets for the other nodes, and (2) as test sets in order to prevent over-fitting phenomenon during training process of the LSTM network. Then, network models will be sent back to the edge for the purpose of \dr~ prediction. As more new devices added to the framework, the prediction error will be decreased as we demonstrated in the next section.

%
\section{Experimental Results}\label{sec:ExpResults}

The performance of proposed real-time reliability analysis was examined for training the data and applying the Deep RACE. This section describes the testing scenarios, the hardware setup, and the experimental results.

\subsection{Experimental training of power transistors}\label{sub:ExpSetup}

On the cloud server, we used Intel Xeon CPU E5-2640 to train the deep LSTM network, where we initially modeled Si-power MOSFETs. The experimental data sets for both training and testing of the power MOSFET (IRF520NPbf) are provided from NASA dataset \cite{celaya2011prognostics}. The Deep RACE predicts a new transistor degradation behavior based on the trained system without any prior knowledge in advance. In our experiment, the Deep RACE is trained to estimate the next 104 samples. For the application with higher window resolutions (i.e. higher output sequence), the network input sequence should also be increased to minimize the prediction error. Table~\ref{table:LSTM_Config} summarizes the deep LSTM network parameters. Based on the network configuration, we have also measured the training time on the cloud server. Table \ref{table:trainingTime} summarizes the training time on the cloud side.

\begin{table}[h]
	\caption{The parameters for LSTM network training}
		\centering 
		\scalebox{1.0}{
			\begin{tabular}{c c c c c}
				\hline
				\hline
				Item & Parameter \ & Description \ & Value
				\\[0.1ex]
				\hline                  
				1 & $k$ & Input size  & 1
				\\
				2 & $\tau$  & Input sequence  & 21
				\\
				3 & $e_{th}$ & Error threshold & $5\times10^{-5}$
				\\
				4 & $n$ & Output sequence   & 104
				\\
				5 & $it_{max}$ & Maximum iterations & 1000	
				\\
				6 & $\epsilon$ & Number of hidden layer & 64  
				\\
				7 & $\ell$ & Number of stacked layer & 4
				\\  
				8 & $m$ & Number of device for training & 4     
				\\	
								\hline			
			\end{tabular}
			\label{table:LSTM_Config}
		}
\end{table}

We used Google TensorFlow framework to implement our stacked LSTM network model. Each LSTM cell is instantiated by calling $tensorflow.contrib.rnn.LSTMCell$ function where the number of ``hidden layer" is passed as an argument to this function. In the next step, an array consists of $LSTMCell$ with the size of ``stacked layer" is generated. Then, the array will be passed to the $tensorflow.contrib.rnn.MultiRNNCell$ function to create the stacked LSTM network. The network unrolling is accomplished through $tensorflow.nn.dynamic\_rnn$ function. We defined Mean Square Error (MSE) (\ref{eq:MSE}) as an objective loss function, and used $tf.train.AdamOptimizer$ method to minimize the function:
\begin{equation}\label{eq:MSE}
MSE=\dfrac{1}{n}\sum_{i=1}^{n}({y}_i - z_i(\theta))^2,
\end{equation}
where $z_i(\theta)$ is the predicted output trajectory from the Deep RACE, and $y_i$ is the actual measurement of the device resistance. 

\begin{table}[h]
	\caption{Data size and average elapsed time for training}
	\centering 
	\scalebox{1.0}{
		\begin{tabular}{c c c}
			\hline
			\hline
			Item & Description \ & Value
			\\[0.1ex]
			\hline                  
			1 &  Training size/Iteration  & $m \times k(\tau+n) = 500$
			\\
			2 &  CPU Cores  & 32
			\\
			3 &  GPUs & \makecell{nVidia Tesla P100, \textit{and} \\nVidia TITAN V}
			\\
			4 &  Elapsed time & 596 Seconds
			\\
			\hline			
		\end{tabular}
		\label{table:trainingTime}
	}
\end{table}

\subsection{Edge Node Hardware Setup}\label{sub:Reliability_Analysis}

We also developed a low-power edge computing system for real-time monitoring and reliability assessment. The edge computing node is based on nVidia TX2 board as the state-of-art embedded SoC with GPU compute units for edge device. As we explained, the inference part of Deep RACE is implemented on the edge, since the cloud is responsible for aggregated training of the proposed structure. 

\begin{figure}[h]
	\centering
	\includegraphics[width=0.3\textwidth,trim= 1 1 1 1,clip]{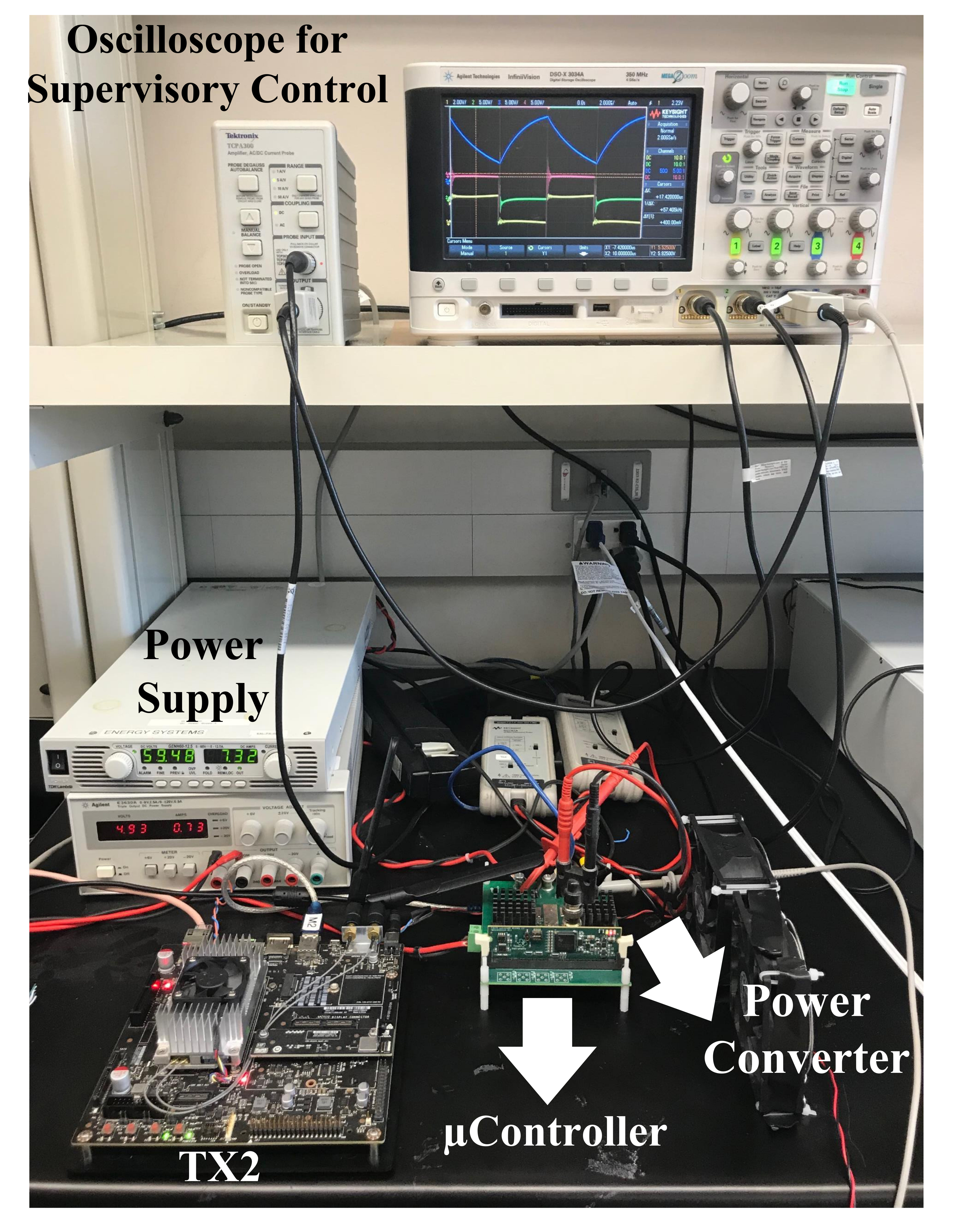}
	\caption{\textbf{Experimental Verification}: The hardware realization of Deep RACE including the high-frequency power converter controller, and the SoC-TX2 for edge computation. The supervisory control is designed for the safety protection.}
	\label{fig:TX2}
\end{figure}

Fig.~\ref{fig:TX2} shows the prototype of Deep RACE hardware realization at the edge. In this system, $\mu$-controller controls the power converter, and the voltage, and current of the power semiconductor are captured and then transfered to the TX2 board for edge analysis. For the safety purpose, the automated supervisory control is designed for data collection from the switching converter and also protects the system operation if the power conversion deviates more than 5\%.

\subsection{Reliability Modeling and Prediction}
We created five different scenarios per each device in order to evaluate the scalability and robustness of Deep RACE. As an example, the trajectory resistance for Dev\#5 is predicted based on learning degradation model from Dev\#1 to Dev\#4. Then, recursively we reinitialized all network models from truncated normal distribution again, and we substituted the other devices to predict a new unknown transistor resistance variations from scratch. Therefore, the Deep RACE is challenged to predict a completely new and unknown device based on aggregating knowledge from other power devices at each scenario. We verified the system characteristics from acquired experimental results in two forms of MSE and error distribution. Table~\ref{table:MSE} shows prediction MSE of the selected devices. While the training process is performed in the cloud, we evaluated the prediction of the device resistance variation at the edge.

\begin{figure}[h]
	\centering
	\subfigure[Dev1]
	{
		\includegraphics[width=0.45\linewidth,trim= 5 1 5 1,clip]{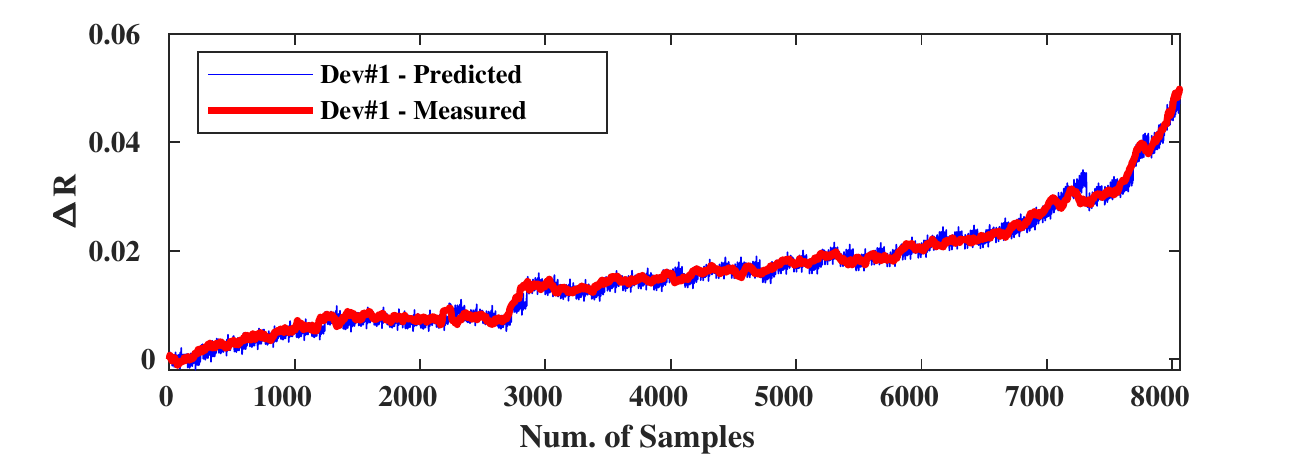}  
		\label{fig:dev9}
	}
	\subfigure[Dev2]
	{
		\includegraphics[width=0.45\linewidth,trim= 5 1 5 1,clip]{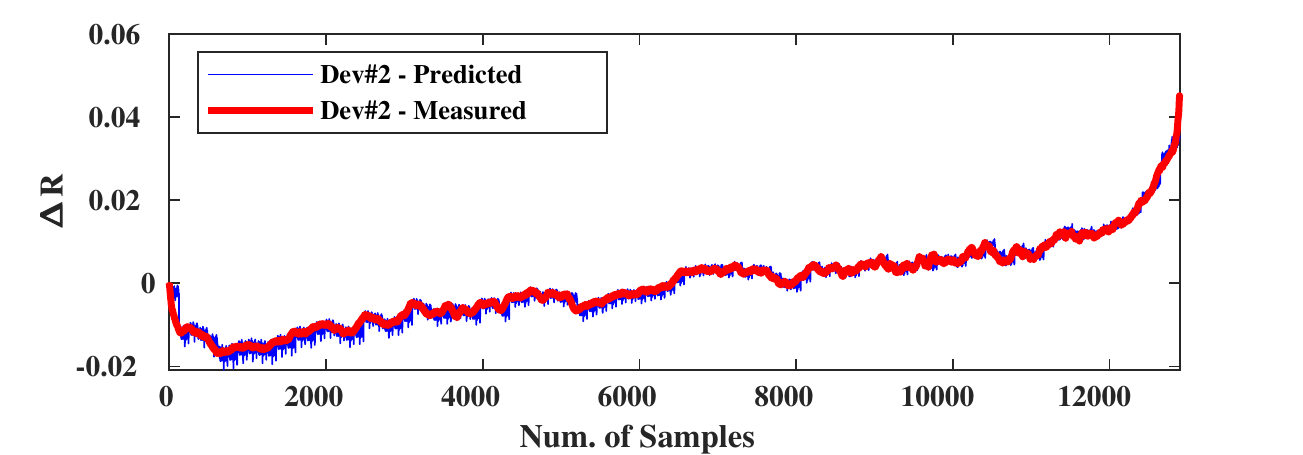}
		\label{fig:dev11}
	}
	\subfigure[Dev3]
	{
		\includegraphics[width=0.45\linewidth,trim= 5 1 5 1,clip]{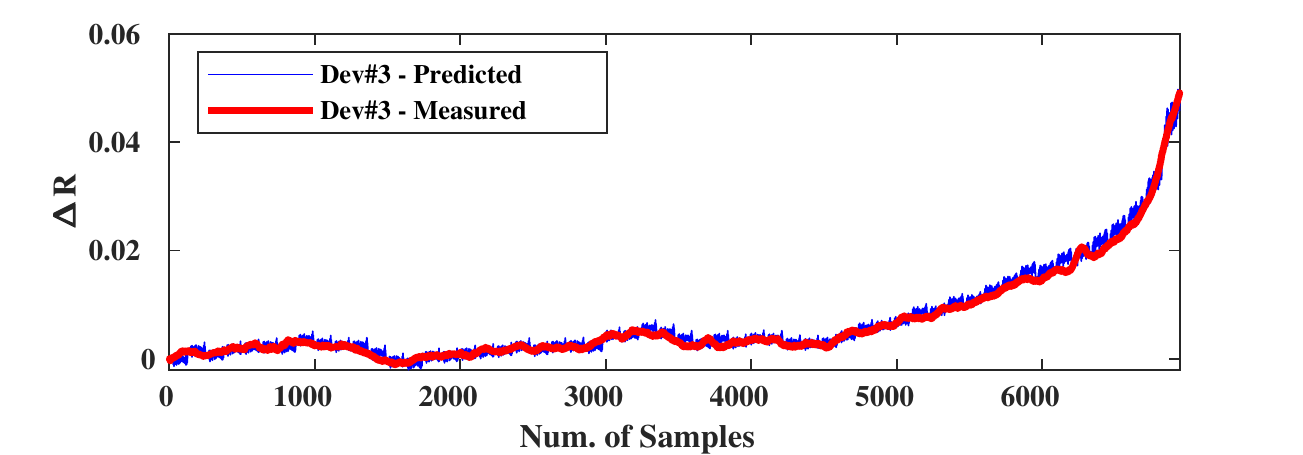}
		\label{fig:dev12}
	}
	\subfigure[Dev4]
	{
		\includegraphics[width=0.45\linewidth,trim= 5 1 5 1,clip]{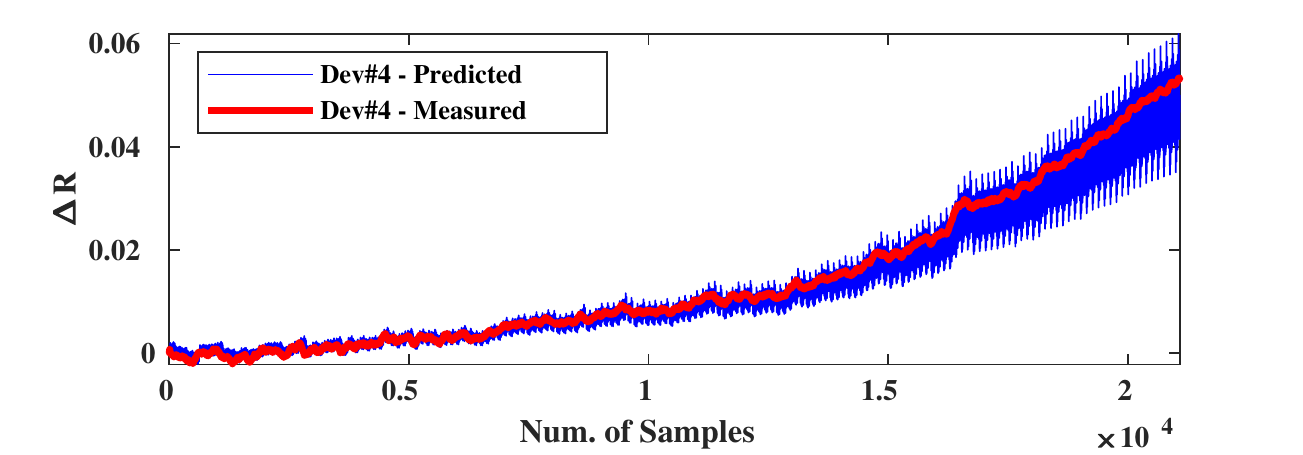}
		\label{fig:dev36}
	}

	\subfigure[Dev5]
	{
		\includegraphics[width=0.45\linewidth,trim= 5 1 5 1,clip]{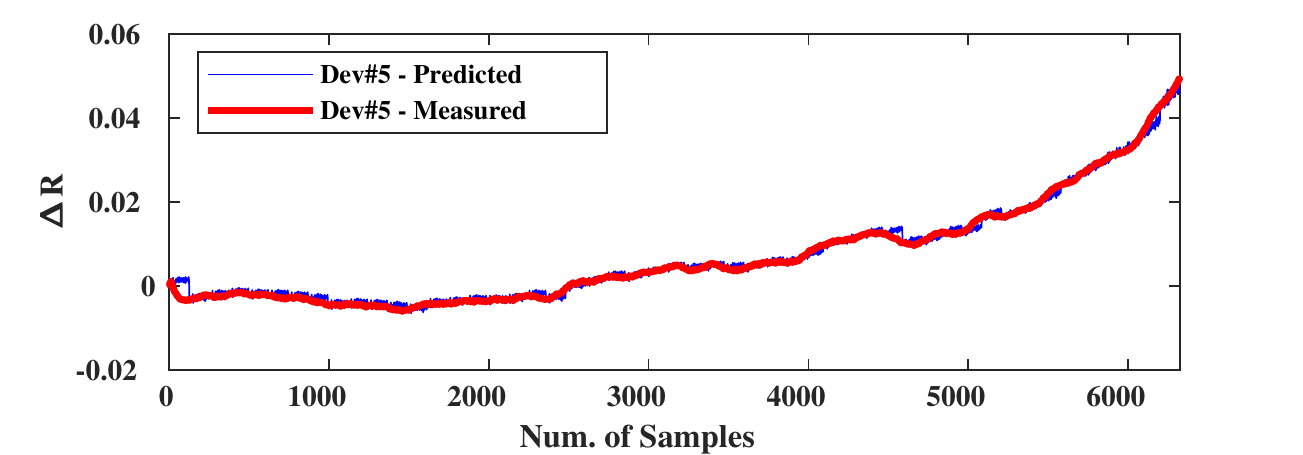}
		\label{fig:dev38}
	}
	\caption{\textbf{Experimental results:} The resistance variation of given five power modules, which were predicted by Deep RACE method.}
	\label{fig:devs}
\end{figure}

Fig.~\ref{fig:devs} illustrates the Deep RACE prediction performance for defined five scenarios and clarifies the scalability of the proposed algorithm. Although the apparatus behavior of each power device degradation looks similar, the microscopic observation of the transistors is different within the same time horizon. For an instance, the trained network for Dev\#4 is expecting an exponential increment in the region of \dr~ $>0.02\Omega$ based on aggregated training from Dev\#1 to Dev\#3, and Dev\#5. This error can be further minimized through a new learning phase on the cloud. 

\begin{table}[h]
	\caption{Prediction error for the power MOSFET transistors.}
	\begin{center}
		\scalebox{1.05}{
			\begin{tabular}{c c c c c c}
				\hline
				\hline
				\text{Devices} &  \text{\#1} & \text{\#2} & \text{\#3}& \text{\#4} & \text{\#5}\\
				\hline
				\text{log(MSE)} & -13.61 & -13.05 & -13.95 & -13.36 & -12.94\\
				\hline
			\end{tabular}
			\label{table:MSE}
			
		}
	\end{center}
\end{table}

\begin{figure}[h]
	\centering
	\includegraphics[width=0.6\linewidth,trim= 20 20 20 20,clip]{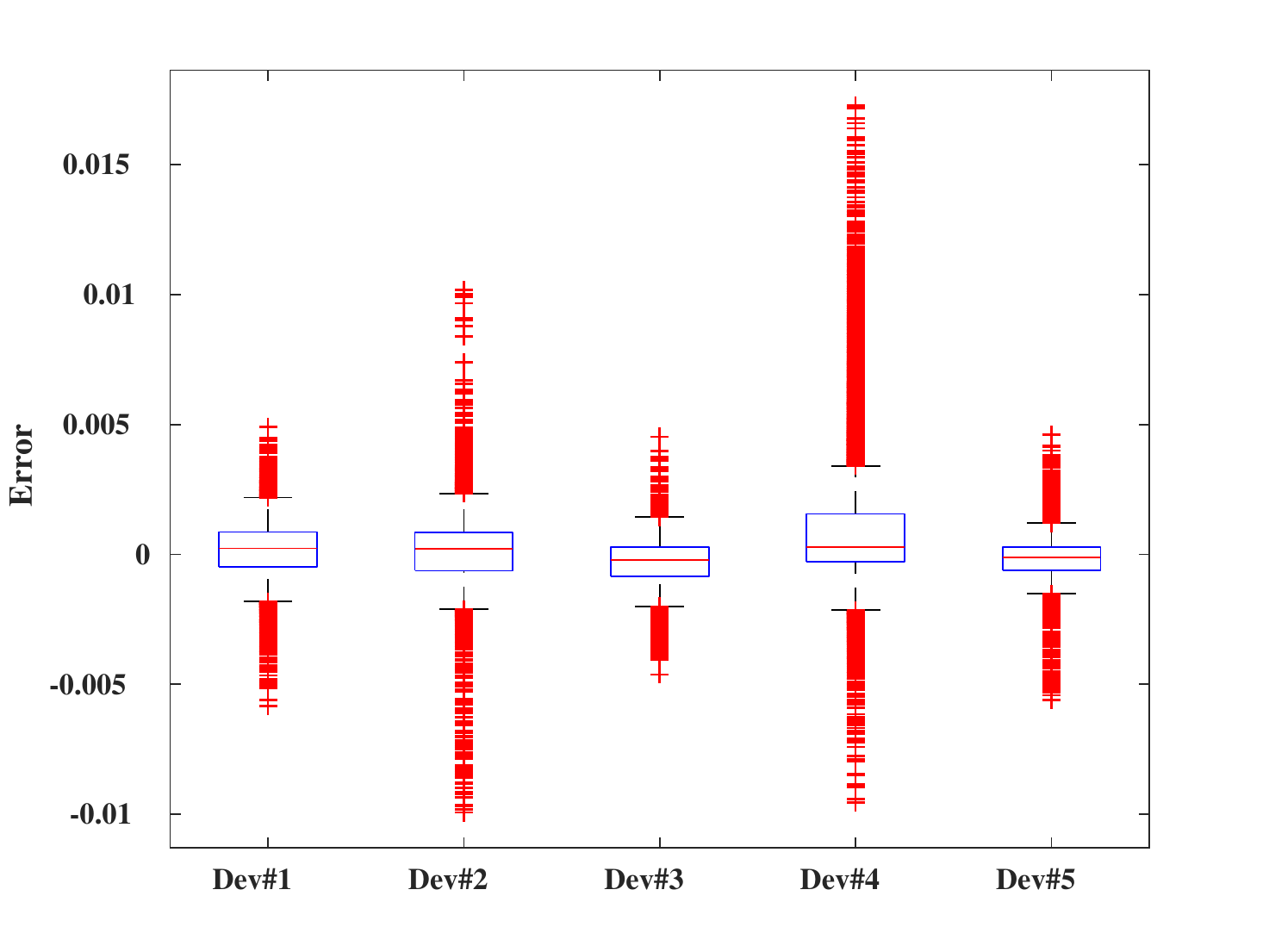}
	\caption{\textbf{Error distribution}: The box-whisker plots of prediction error for five power modules.}
	\label{fig:error}
\end{figure}

We also extracted the error distribution for five predicted devices by using: 
\begin{equation}\label{eq:Absolute_error_1}
Error_{diff}=({y}_i - z_i(\theta)),
\end{equation}
where $z_i(\theta)$ is the predicted output trajectory from the Deep RACE, and $y_i$ is the actual measurement of the device resistance (\dr~in our model). Fig.~\ref{fig:error} depicts that the average maximum error caused by Deep RACE method is less than 0.9\%.  

\begin{figure}[h]
	\centering
	\includegraphics[width=0.6\linewidth,trim= 10 5 20 20,clip]{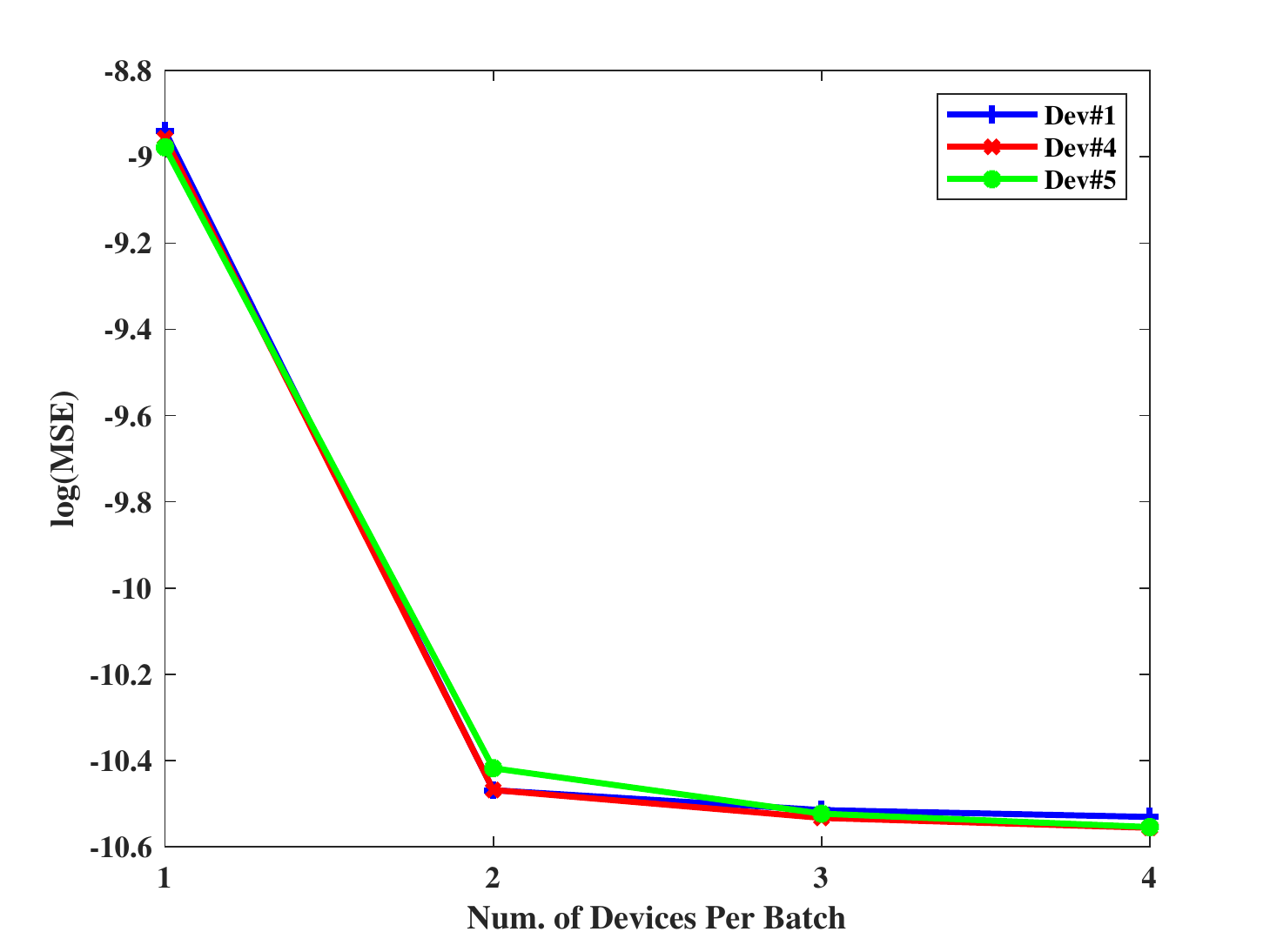}
	\caption{\textbf{Aggregated training}: The \dr~ prediction error is decreased exponentially by increasing the training device per each batch. This aggrigated training will help the Deep RACE to generalize the different transistor degradation behavior.}
	\label{fig:errorPerDevice}
\end{figure}

We extended our experiment to analyze the effect of training aggregation and scalability of multiple device data on accuracy of \dr~prediction. In this case, we increased the number of devices per each batch during the training phase. Fig.~\ref{fig:errorPerDevice} shows that MSE decreases with an exponential rate by increasing the number of devices in training batch. For each training set, we ran 1000 Monte-Carlo test for \dr~predictiopn of three different devices, and then the average of whole test sets is picked. These results indicate that our proposed approach can improve the prediction accuracy exponentially by increasing the edge node and power transistors through the life time of the system by accumulating knowledge about different device degradation during its usage, which demonstrates the scalability of our approach.


The region of \dr~$< 0.02\Omega$ has linear behavior, therefore, the classical approaches (such as Kalman Filter, and Particle Filter) can predict the health state with higher accuracy; however, the prediction error increases after that for these methods. Since it is very crucial to detect the MOSFET resistance variation when \dr~ $\approx 0.05\Omega$, we calculated the average of error at the detection point(=${\Delta R}_{5\%}$) for these methods by using  (\ref{eq:erro-comparision}):

\begin{equation}\label{eq:erro-comparision}
{Error_{{\Delta R}_{5\%}}}=\dfrac{100}{m}\sum_{i=1}^{m}{|{{\Delta R}_{ds(on)_{mt_{5\%}}} - (0.05)_{mt_{5\%}}}|\over{(0.05)_{mt_{5\%}}}},
\end{equation}
where $m$ is the number of devices, ${\Delta R}_{ds(on)}$ is predicted value, and $t_{5\%}$ is the time when the first sensed value is 0.05$\Omega$. The $Error_{{\Delta R}_{5\%}}$ results are summarized in Table~\ref{table:modelComp}. Our experiments indicate that the Deep RACE reduces the miss-prediction error at $0.05\Omega$ by about 1.98x, 1.77x  compared to Kalman Filter and Particle Filter, respectively.

\begin{table}[h]
	\caption{The comparison of the absolute average error of Deep RACE with conventional methods}
	\centering 
\scalebox{1.05}{
	\begin{tabular}{c c c c c}
		\hline
		\hline
				\text{\shortstack{~ \\ Method \\~ \\~ \\~}} &  \text{\shortstack{Kalman \\Filter \\~\cite{dusmez2016remaining}}} & \text{\shortstack{Particle \\Filter\\ ~\cite{celaya2011prognostics}}}& \text{\shortstack{Deep RACE \\~\\~}} \\
				\hline
				\text{\shortstack{~\\ Miss-prediction \\Error}} & 17.75\% & 15.85\% & 8.93\% \\
				\hline
			\end{tabular}
		\label{table:modelComp}
	}
\end{table}

\subsection{Power consumption and processing time analysis}\label{sub:effectOfTrainingSet}
We also evaluated the power consumption and delay of the inference part of the network on embedded TX2 board. Table~\ref{table:TX2} summarizes the specification of embedded SoC. Since the TX2 has an embedded GPU, We considers two different scenarios to analyze the performance of Deep RACE. At first scenario, we set the tensorflow configuration to \verb|device_count = {'GPU': 0}|, where no computation carried out at the embedded GPU, and in the second approach we made it ON. For the matrix size of 125 (input sequence + output sequence), it was observed that the CPU processed 3.2x faster than GPU for one device prediction. This performance degradation is because of data copying between CPU and GPU memory region -- Note the DDR power consumption is higher for \verb|'GPU': 1| scenario. In the other word, the amount of data is not enough for GPU to overlap the delay between data computation and movement. Increasing the number of devices that should be predicted per each edge node or increasing the prediction window resolution (\emph{output sequence}) improves the performance for GPU since it carries out more computation than CPU per each data set. Table~\ref{table:TX2_POWER} summarizes the delay and power dissipation for two different cases.

\begin{table}[h]
	\centering 
	\caption{nVidia TX2 Embedded Module Specification}
	\begin{center}
		\scalebox{1.05}{
			\begin{tabular}{c c c }
				\hline
				\hline
				\text{\makecell{CPU}}&\text{\makecell{GPU}} &  \text{\makecell{DDR}}\\
				\hline
				\makecell{Quad Cortex-A57 \\@ 2GHz +\\
					Dual Denver2 @ 2GHz} & \makecell{256-core \\Pascal\\ @ 1300MHz} & \makecell{8GB LPDDR4\\ @ 1866MHz}\\
				\hline
			\end{tabular}
			\label{table:TX2}
		}
	\end{center}
\end{table}

\begin{table}[h]
	\renewcommand{\arraystretch}{1.00}
	\centering 
	\caption{TX2 embedded board power consumption.}
	
	\scalebox{0.99}{
		\begin{tabular}{ c c c c||c c c }
			\hline
			\hline
			\multicolumn{1}{c }{} &\multicolumn{3}{c||}{GPU: OFF (0)} & \multicolumn{3}{c }{GPU: ON (1)}				\\[0.1ex]
			\hline
			\text{Module} &  \text{CPU} & \text{DDR} & \text{CPU+DDR}& \text{GPU} & \text{DDR}& \text{GPU+DDR}\\
			\hline
			\text{\shortstack{Power \\ (W)}} & 1.07 & 0.80 & 1.87 & 0.166 & 0.90 & 1.06\\
			\hline
			\text{\shortstack{Delay \\ (ms)}}& \multicolumn{3}{c||}{26} & \multicolumn{3}{c }{85}\\
			\hline
		\end{tabular}
		\label{table:TX2_POWER}
	}
	
\end{table}

%
\section{Conclusion and Future Work}\label{sec:Conclusion}
This paper proposed a new solution as a collection of deep learning, edge, and cloud computing technologies to enable real-time high accuracy reliability modeling of high-frequency MOSFETs power converter devices. The proposed deep learning algorithm is based on LSTM algorithmic constructs for accumulating the degradation knowledge of different power MOSFET devices on the cloud server, and real-time inference at the edge. For the experimented results, we developed an entire integrated system of Deep RACE, including an embedded system system-on-chip implementation on nVidia SoC-TX2. The results demonstrated the real-time convergence of the system with about $8.9\%$ miss prediction, with $26ms$ processing time. 

In a broader perspective, the proposed research will have a fundamental contribution in the engineering of semiconductor devices and information processing by bringing recent advances in deep learning and edge computing for real-time predictive maintenance of emerging semiconductor devices. In this context, Deep RACE sets to move beyond mainstream device modeling and traditional reliability analysis (i.e. Weibull distributions, mean-time-to-failure, etc.) and looking to more applicable and accurate analytical tools through introducing advanced sensing solutions and combining it with cutting-edge deep learning techniques.

%
\section*{Acknowledgment}

The authors would like to thank Energy Production and Infrastructure Center (EPIC), and ECE Department at the University of North Carolina at Charlotte.




%


\bibliographystyle{IEEEtran}
\scriptsize
\bibliography{references}

\end{document}